\documentclass[11pt,a4paper]{article}
\usepackage[hyperref]{emnlp2018}
\usepackage{times}
\usepackage{latexsym}
\usepackage{color}
\usepackage{graphicx}
\usepackage{multirow}

\usepackage{url}

\aclfinalcopy 

\title{Rapid Adaptation of Neural Machine Translation to New Languages}

\author{Graham Neubig, Junjie Hu \\
  Language Technologies Institute, Carnegie Mellon University \\
  {\tt \{gneubig,junjieh\}@cs.cmu.edu}}

\date{}

\begin{document}
\maketitle
\begin{abstract}
  This paper examines the problem of adapting neural machine translation systems to new, low-resourced languages (LRLs) as \emph{effectively} and \emph{rapidly} as possible.
  We propose methods based on starting with massively multilingual ``seed models'', which can be trained ahead-of-time, and then continuing training on data related to the LRL.
  We contrast a number of strategies, leading to a novel, simple, yet effective method of ``similar-language regularization'', where we jointly train on both a LRL of interest and a similar high-resourced language to prevent over-fitting to small LRL data.
  Experiments demonstrate that massively multilingual models, even without any explicit adaptation, are surprisingly effective, achieving BLEU scores of up to 15.5 with \emph{no} data from the LRL, and that the proposed similar-language regularization method improves over other adaptation methods by 1.7 BLEU points average over 4 LRL settings.\footnote{Code to reproduce experiments at \url{https://github.com/neubig/rapid-adaptation}}
\end{abstract}

\section{Introduction}
\label{sec:intro}

When disaster strikes, news and social media are invaluable sources of information, allowing humanitarian organizations to rapidly mitigate crisis situations and save lives \cite{vieweg10situational,anpinlp11ijcnlp,starbird12ontheground}.
However, language barriers looms large over these efforts, especially when disasters occur in parts of the world that use less common languages.
In these cases, machine translation (MT) technology can be a valuable tool, with one widely-heralded success story being the deployment of Haitian Creole-to-English translation systems during the earthquakes in Haiti \cite{lewis10haitian,munro10crowdsourced}.

However, data-driven MT systems, particularly neural machine translation (NMT; \newcite{kalchbrenner13rnntm,bahdanau15alignandtranslate}), require large amounts of training data, and creating high-quality systems in low-resource languages (LRLs) is a difficult challenge where research efforts have just begun \cite{gu2018universal}.
Another hurdle, which to our knowledge has not been covered in previous research,
is the \emph{time} it takes to create such a system.
In a crisis situation, time is of the essence, and systems that require days or weeks of training will not be desirable or even feasible.

In this paper we focus on the question: \emph{how can we create MT systems for new language pairs as accurately as possible, and as quickly as possible?}
To examine this question we propose NMT methods at the intersection of cross-lingual transfer learning \cite{zoph2016transfer} and multilingual training \cite{johnson16multilingual}, two paradigms that, to our knowledge, have not been used together in previous work.
Our methods, laid out in \S\ref{sec:paradigms} follow the process of training a seed model on a large number of languages, then fine-tuning the model to improve its performance on the language of interest.
We propose a novel method of \emph{similar-language regularization} (SLR) where training data from a second similar languages is used to help prevent over-fitting to the small LRL dataset.

In the experiments in \S\ref{sec:exp}, we attempt to answer two questions: 
(1) Which method of creating multilingual systems and adapting them to an LRL is the most effective way to increase accuracy?
(2) How can we create the strongest system possible with a bare minimum of training time?
The results are sometimes surprising -- we first find that a single monolithic model trained on 57 languages can achieve BLEU scores as high as 15.5 with \emph{no} training data in the new source language whatsoever.
In addition, the proposed method starting with a universal model then fine-tuning with the SLR proves most effective, achieving gains of 1.7 BLEU points averaged over several language pairs compared to previous methods adapting to only the LRL.

\section{Training Paradigms}
\label{sec:paradigms}

In this paper, we consider the setting where we have a \emph{source LRL} of interest, and we want to translate into English.%
\footnote{Translation into LRLs, is a challenging and interesting problem in it's own right, but beyond the scope of the paper.}
All of our adaptation methods are based on first training on larger data including other languages, then fine-tuning the model to be specifically tailored to the LRL.
We first discuss a few multilingual training paradigms from previous literature (\S\ref{sec:modeling}), then discuss our proposed adaptation methods (\S\ref{sec:adaptation}).

\subsection{Multilingual Modeling Methods}
\label{sec:modeling}

We use three varieties of multilingual training:

\noindent \textbf{Single-source modeling (``Sing.'')}
is the first method, using only parallel data between the LRL of interest and English.
This method is straightforward and the resulting model will be most highly tailored to the final test language pair, but the method also has the obvious disadvantage that training data is very sparse.

\noindent \textbf{Bi-source modeling (``Bi'')}
trains an MT system with two source languages: one LRL that we would like to translate from, and a second highly related high-resource language (HRL): the \emph{helper} source language.%
\footnote{``Related'' could mean different things: typologically related or having high lexical overlap. In our experiments our LRLs are all selected to have an helper that is highly similar in both aspects, but choosing an appropriate helper when this is not the case is an interesting problem for future work.}
This method is inspired by \newcite{johnson16multilingual}, who examine multilingual translation models to/from English and two highly related languages such as Spanish/Portuguese or Japanese/Korean.
The advantage of this method is that it allows the LRL to learn from a highly similar helper, potentially increasing accuracy.

\noindent \textbf{All-source modeling (``All'')}
trains not only on a couple source languages, but instead creates a universal model on all of the languages that we have at our disposal.
In our experiments (\S\ref{sec:exp:setup}) this entails training systems on 58 source languages, to our knowledge the largest reported in NMT experiments.\footnote{In contrast to \newcite{gu2018universal}, who train on 10 languages. \newcite{malaviya17emnlp,tiedemann2018emerging} train NMT on over 1,000 languages, but only as a feature extractor for downstream tasks; MT accuracy itself is not evaluated.}
This paradigm allows us to train a single model that has wide coverage of vocabulary and syntax of a large number of languages, but also has the drawback in that a single model must be able to express information about all the languages in the training set within its limited parameter budget.
Thus, it is reasonable to expect that this model may achieve worse accuracy than a model created specifically to handle a particular source language.

In the following, we will consider adaptation methods that focus on tailoring a more general model (i.e. bi-source or universal) to a more specific model (i.e. single-source or bi-source).

\subsection{Adaptation to New Languages}
\label{sec:adaptation}

As noted in the introduction, there are two major requirements: the \emph{accuracy} of the system is important and the \emph{training time required} from when we learn of a need for translation to when we can first start producing adequate results.
Throughout the discussion, we will compare various adaptation paradigms with respect to these two aspects.

\subsubsection{Adaptation by Fine-tuning}
\label{sec:paradigms:finetuning}

Our first adaptation method, inspired by \newcite{zoph2016transfer} is based on fine-tuning to the source language of interest. Within our experiments, we will test this setting, but also make two distinctions between the types of adaptation:

\noindent \textbf{Seed Model Variety:} \newcite{zoph2016transfer} performed experiments taking a bilingual system trained on a different language (e.g. French) and adapting it to a new LRL (e.g. Uzbek).
We can also take universal model and adapt it to the new language, a setting that we examine (to our knowledge, for the first time) in this work.

\noindent \textbf{Warm vs. Cold Start:}
Another contrast is whether we have training data for the LRL of interest while training the original system, or whether we only receive training data \emph{after} the original model has already been trained.
We call the former \emph{warm start}, and the latter \emph{cold start}.
Intuitively, we expect warm-start training to perform better, as having access to the LRL of interest during the training of the original model will ensure that it can handle the LRL to some extent.
However, the cold-start scenario is also of interest: we may want to spend large amounts of time training a strong model, then quickly adapt to a new language that we have never seen before in our training data as data becomes available.
For the cold-start models, we start with a model that is only trained on the HRL similar to the LRL (Bi$^-$), or a model trained on all languages \emph{but} the LRL (All$^-$).


\subsubsection{Similar-Language Regularization}
\label{sec:paradigms:similar}

One problem with adapting to a small amount of data in the target language is that it will be very easy for the model to over-fit to the small training set.
To alleviate this problem, we propose a method of \emph{similar language regularization}: while training to adapt to the language of interest, we also add some data from another similar HRL that has sufficient resources to help prevent over-fitting.
We do this in two ways:

\noindent \textbf{Corpus Concatenation:}
Simply concatenate the data from the two corpora, so that we have a small amount of data in the LRL, and a large amount of data in the similar HRL.

\noindent \textbf{Balanced Sampling:}
Every time we select a mini-batch to do training, we either sample it from the LRL, or from the HRL according to a fixed ratio.
We try different sampling strategies, including sampling with a 1-to-1 ratio, 1-to-2 ratio, and 1-to-4 ratio for the LRL and HRL respectively.

\section{Experiments}
\label{sec:exp}

\subsection{Experimental Setup}
\label{sec:exp:setup}

We perform experiments on the 58-language-to-English TED corpus \cite{qi18naacl}, which is ideal for our purposes because it has a wide variety of languages over several language families, some high-resourced and some low-resourced.
Like \newcite{qi18naacl}, we experiment with Azerbaijani (aze), Belarusian (bel), and Galician (glg) to English, and also additionally add Slovak (slk), a slightly higher resourced language, for contrast.
These languages are all paired with a similar HRL: Turkish (tur), Russian (rus), Portuguese (por), and Czech (ces) respectively.
Data sizes are shown in Table~\ref{tab:datasize}.

\begin{table}
\centering
\small
\begin{tabular}{l|rrr||l|r}
LRL & train & dev   & test  & HRL & train \\ \hline
aze & 5.94k & 671   & 903   & tur & 182k \\
bel & 4.51k & 248   & 664   & rus & 208k \\
glg & 10.0k & 682   & 1,007 & por & 185k \\
slk & 61.5k & 2,271 & 2,445 & ces & 103k \\
\end{tabular}
  \vspace{-2mm}
\caption{Data sizes in sentences for LRL/HRL pairs \label{tab:datasize}}
  \vspace{-5mm}
\end{table}

Models are implemented using \texttt{xnmt} \cite{neubig18xnmt}, commit \texttt{8173b1f}, and start with the recipe for training on IWSLT TED\footnote{Found in \texttt{examples/stanford-iwslt/}}.
The model consists of an attentional neural machine translation model \cite{bahdanau15alignandtranslate}, using bi-directional LSTM encoders, 128-dimensional word embeddings, 512-dimensional hidden states, and a standard LSTM-based decoder.

Following standard practice \cite{sennrich16bpe,denkowski17nmt}, we break low-frequency words into subwords using the \texttt{sentencepiece} toolkit.\footnote{\url{https://github.com/google/sentencepiece}, using the \texttt{unigram} training setting.}
There are two alternatives for creating subword units: \emph{jointly} learning subwords over all source language, or \emph{separately} learning subwords for each source language, then taking the union of all the subword vocabularies as the vocabulary for the multilingual model.
Previous work on multilingual training has preferred the former \cite{nguyen-chiang:2017:I17-2}, but in this paper we use the latter for two reasons: (1) because data in the LRL will not affect the subword units from the other languages, in the cold-start scenario we can postpone creation of subword units for the LRL until directly before we start training on the LRL itself, and (2) we need not be concerned with the LRL being ``overwhelmed'' by the higher-resourced languages when calculating statistics used in the creation of subword units, because all languages get an equal share.%
\footnote{
Preliminary experiments found both comparable: with scores of 20.1 and 19.4 for separate and joint respectively.
}
In the experiments, we use a subword vocabulary of 8,000 for each language.

We also compare with two additional baselines: phrase-based MT implemented in \texttt{Moses},\footnote{\url{http://statmt.org/moses}} and unsupervised NMT implemented in \texttt{undreamt}.\footnote{\url{https://github.com/artetxem/undreamt}}
\texttt{Moses} is trained on the bilingual data only (training multilingually reduced average accuracy), and \texttt{undreamt} is trained on all monolingual data available for the LRL and English.

\subsection{Experimental Results}
\label{sec:exp:results}

Table \ref{tab:main} shows our main translation results, with warm-start scenarios in the upper half and cold-start scenarios in the lower half.

\paragraph{Does Multilingual Training Help?} To answer this question, we can compare the warm-start Sing., Bi, and All settings, and find that the answer is a resounding yes, gains of 7-13 BLEU points are obtained by going from single-source to bi-source or all-source training, corroborating previous work \cite{gu2018universal}.
Bi-source models tend to perform slightly better than all-source models, indicating that given identical parameter capacity, training on a highly resourced language is effective.
Comparing with the phrase-based baseline, as noted by \newcite{koehn2017six} NMT tends to underperform on low-resource settings when trained only on the data available for these languages.
However, multilingual training of any variety quickly remedies this issue; all outperform phrase-based handily.

\begin{table}
\resizebox{\columnwidth}{!}{
\begin{tabular}{ll|rrrr|r}
 & Strategy & aze/tur & bel/rus & glg/por & slk/ces & Avg. \\ \hline \hline
 \multicolumn{2}{l|}{Phrase-based}     & 5.9 & 10.5 & 22.3 & 23.0 & 15.4 \\
 \multicolumn{2}{l|}{Unsupervised NMT} & 0.0 & 0.3 & 0.4 & 0.0 & 0.2 \\ \hline
\multirow{ 9 }{*}{\rotatebox{90}{ Warm Start }} & Sing. & 2.7 & 2.8 & 16.2 & 24.0 & 11.4\\
 & Bi & 10.9 & 15.8 & 27.3 & 26.5 & 20.1\\
 & All & 9.7 & 16.7 & 26.5 & 25.0 & 19.5\\
 & Bi $\rightarrow$Sing. & 11.4 & 16.3 & 27.5 & 27.1 & 20.6\\
 & All$\rightarrow$Sing. & 10.1 & 17.5 & 28.2 & 27.4 & 20.8\\
 & All$\rightarrow$Bi & \textbf{11.7} & \textbf{18.3} & 28.8 & 28.2 & \textbf{21.8}\\
 & All$\rightarrow$Bi 1-1 & 10.2 & 18.3 & 28.8 & \textbf{28.3} & 21.4\\
 & All$\rightarrow$Bi 1-2 & 11.0 & 17.5 & \textbf{29.1} & 28.2 & 21.4\\
 & All$\rightarrow$Bi 1-4 & 11.1 & 17.9 & 28.5 & 27.9 & 21.3\\ \hline
\multirow{ 8 }{*}{\rotatebox{90}{ Cold Start }} & Bi$^-$ & 3.8 & 2.5 & 8.6 & 5.4 & 5.1\\
 & All$^-$ & 3.7 & 3.5 & 15.5 & 7.3 & 7.5\\
 & Bi$^-$ $\rightarrow$Sing. & 8.7 & 11.8 & 25.4 & 26.8 & 18.2\\
 & All$^-$$\rightarrow$Sing. & 8.8 & 15.3 & 26.5 & 27.6 & 19.5\\
 & All$^-$$\rightarrow$Bi & 10.7 & \textbf{17.4} & 28.4 & 28.0 & \textbf{21.2}\\
 & All$^-$$\rightarrow$Bi 1-1 & 10.5 & 16.0 & 28.0 & \textbf{28.2} & 20.7\\
 & All$^-$$\rightarrow$Bi 1-2 & 10.7 & 17.1 & 28.3 & 27.9 & 21.0\\
 & All$^-$$\rightarrow$Bi 1-4 & \textbf{11.0} & 17.4 & \textbf{28.4} & 27.6 & 21.1\\
\end{tabular}
}
  \vspace{-2mm}
\caption{BLEU for single-source (Sing.), bi-source (Bi), and all-source universal (All) models, with adapted counterparts. 1-1, 1-2, 1-4 indicate balanced sampling from \S\ref{sec:adaptation}. Bold indicates highest score.\label{tab:main}}
  \vspace{-5mm}
\end{table}

More interestingly, examining the cold-start results, we can see that even systems with \emph{no} data in the target language are able to achieve non-trivial accuracies, up to 15.5 BLEU on glg-eng.
Interestingly, in the cold-start scenario, the All$^-$ model bests the Bi$^-$ model, indicating that massively multilingual training is more useful in this setting. 
In contrast, the unsupervised NMT model struggles, achieving a BLEU score of around 0 for all language pairs -- this is because unsupervised NMT requires high-quality monolingual embeddings from the same distribution, which can be trained easily in English, but are not available in the low-resource languages we are considering.

\noindent \textbf{Does Adaptation Help?}
Regarding adaptation, we can first observe that regardless of the original model and method for adaptation, adaptation is helpful, particularly (and unsurprisingly) in the cold-start case.
When adapting directly to only the target language (``$\rightarrow$Sing.''), adapting from the massively multilingual model performs better, indicating that information about all input languages is better than just a single language.
Next, comparing with our proposed method of adding similar language regularization (``$\rightarrow$Bi''), we can see that this helps significantly over adapting directly to the LRL, particularly in the cold-start case where we can observe gains of up to 1.7 BLEU points.
Finally, in our data setting, corpus concatenation outperforms balanced sampling in both the cold-start and warm-start scenarios.

\begin{figure}[t]
  \centering
  \includegraphics[width=0.46\textwidth]{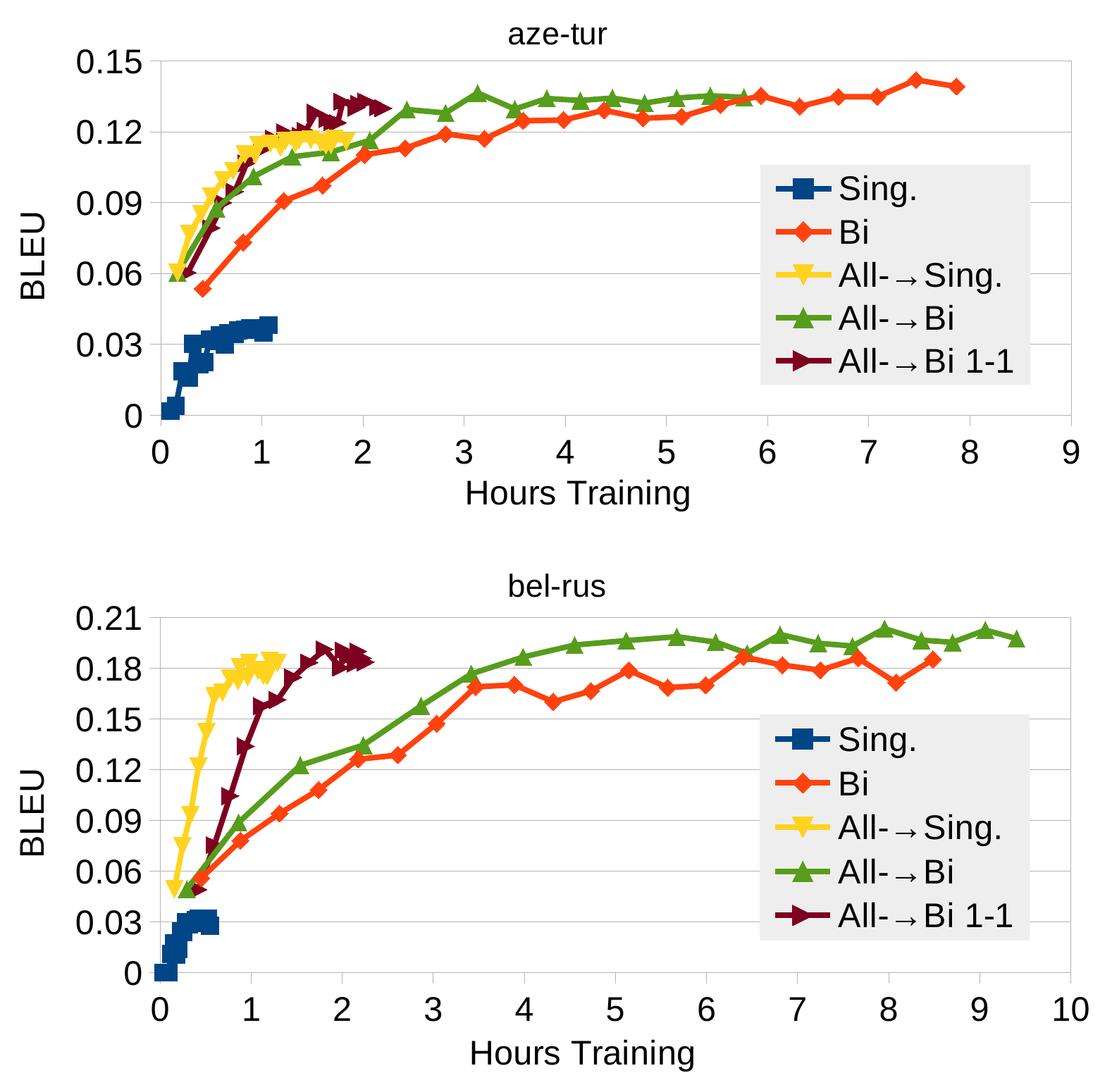}
  \vspace{-2mm}
  \caption{Example of adaptation on the aze-eng and bel-eng development sets}
  \vspace{-5mm}
  \label{fig:adaptation}
\end{figure}

\noindent \textbf{How Can We Adapt Most Efficiently?}
Finally, we revisit adapting to new languages efficiently, with Figure~\ref{fig:adaptation} showing BLEU vs. hours training for the aze/tur and bel/rus source language pairs (others were similar).
We can see that in all cases the cold-start models (All$^-\rightarrow$) either outperform or are comparable in final accuracy to the from-scratch single-source and bi-source models.
In addition, all of the adapted models converge faster than the bi-source from-scratch trained models, indicating that adapting from seed models is a good strategy for rapid construction of MT systems in new languages.
Comparing the cold-start adaptation strategies, we can see that in general, the higher the density of target language training data, the faster the training converges to a solution, but the worse the final solution is.
This suggests that there is a speed/accuracy tradeoff in the amount of similar language regularization we apply during fine-tuning.

\section{Related Work}
\label{sec:related}

While adapting MT systems to new languages is a long-standing challenge 
\cite{schultz2006challenges,jabaian2013comparison}, multilingual NMT is highly promising in its ability to abstract across language boundaries \cite{firat16multiway,ha2016toward,johnson16multilingual}.
Results on multilingual training for low-resource translation \cite{gu2018universal,qi18naacl} further demonstrates this potential, although these works do not consider adaptation to new languages, the main focus of our work.
Notably, we did not examine partial freezing of parameters, another method proven useful for cross-lingual adaptation \cite{zoph2016transfer}; this is orthogonal to our multi-lingual training approach but the two methods could potentially be combined.
Finally, unsupervised NMT approaches
\cite{artetxe2017unsupervised,lample2018phrase,lample2017unsupervised}
require no parallel data, but rest on strong assumptions about high-quality comparable monolingual data.
As we show, when this assumption breaks down these methods fail to function, while our cold-start methods achieve non-trivial accuracies even with no monolingual data.

\section{Conclusion}
\label{sec:conclusion}

This paper examined methods to rapidly adapt MT systems to new languages by fine-tuning.
In both warm-start and cold-start scenarios, the best results were obtained by adapting a pre-trained universal model to the low-resource language while regularizing with similar languages.

\section*{Acknowledgements}

The authors thank Jaime Carbonell, Xinyi Wang, Rebecca Knowles, Arya McCarthy, and anonymous reviewers for their constructive comments on this paper.

This work is sponsored by Defense Advanced Research Projects Agency Information Innovation Office (I2O). Program: Low Resource Languages for Emergent Incidents (LORELEI). Issued by DARPA/I2O under Contract No. HR0011-15-C0114. The views and conclusions contained in this document are those of the authors and should not be interpreted as representing the official policies, either expressed or implied, of the U.S. Government. The U.S. Government is authorized to reproduce and distribute reprints for Government purposes notwithstanding any copyright notation here on.

\bibliography{myabbrv,gneubig}

\begin{thebibliography}{25}
\expandafter\ifx\csname natexlab\endcsname\relax\def\natexlab#1{#1}\fi

\bibitem[{Artetxe et~al.(2017)Artetxe, Labaka, Agirre, and
  Cho}]{artetxe2017unsupervised}
Mikel Artetxe, Gorka Labaka, Eneko Agirre, and Kyunghyun Cho. 2017.
\newblock Unsupervised neural machine translation.
\newblock \emph{arXiv preprint arXiv:1710.11041}.

\bibitem[{Bahdanau et~al.(2015)Bahdanau, Cho, and
  Bengio}]{bahdanau15alignandtranslate}
Dzmitry Bahdanau, Kyunghyun Cho, and Yoshua Bengio. 2015.
\newblock Neural machine translation by jointly learning to align and
  translate.
\newblock In \emph{Proc. ICLR}.

\bibitem[{Denkowski and Neubig(2017)}]{denkowski17nmt}
Michael Denkowski and Graham Neubig. 2017.
\newblock Stronger baselines for trustable results in neural machine
  translation.
\newblock In \emph{Proc. WNMT}.

\bibitem[{Firat et~al.(2016)Firat, Cho, and Bengio}]{firat16multiway}
Orhan Firat, Kyunghyun Cho, and Yoshua Bengio. 2016.
\newblock Multi-way, multilingual neural machine translation with a shared
  attention mechanism.
\newblock In \emph{Proc. NAACL}, pages 866--875.

\bibitem[{Gu et~al.(2018)Gu, Hassan, Devlin, and Li}]{gu2018universal}
Jiatao Gu, Hany Hassan, Jacob Devlin, and Victor~OK Li. 2018.
\newblock Universal neural machine translation for extremely low resource
  languages.
\newblock \emph{Proc. NAACL}.

\bibitem[{Ha et~al.(2016)Ha, Niehues, and Waibel}]{ha2016toward}
Thanh-Le Ha, Jan Niehues, and Alexander Waibel. 2016.
\newblock Toward multilingual neural machine translation with universal encoder
  and decoder.
\newblock \emph{arXiv preprint arXiv:1611.04798}.

\bibitem[{Jabaian et~al.(2013)Jabaian, Besacier, and
  Lefevre}]{jabaian2013comparison}
Bassam Jabaian, Laurent Besacier, and Fabrice Lefevre. 2013.
\newblock Comparison and combination of lightly supervised approaches for
  language portability of a spoken language understanding system.
\newblock \emph{IEEE Transactions on Audio, Speech, and Language Processing},
  21(3):636--648.

\bibitem[{Johnson et~al.(2016)}]{johnson16multilingual}
Melvin Johnson et~al. 2016.
\newblock Google's multilingual neural machine translation system: Enabling
  zero-shot translation.
\newblock \emph{TACL}.

\bibitem[{Kalchbrenner and Blunsom(2013)}]{kalchbrenner13rnntm}
Nal Kalchbrenner and Phil Blunsom. 2013.
\newblock Recurrent continuous translation models.
\newblock In \emph{Proc. EMNLP}, pages 1700--1709.

\bibitem[{Koehn and Knowles(2017)}]{koehn2017six}
Philipp Koehn and Rebecca Knowles. 2017.
\newblock Six challenges for neural machine translation.
\newblock \emph{Proc. WNMT}.

\bibitem[{Lample et~al.(2017)Lample, Denoyer, and
  Ranzato}]{lample2017unsupervised}
Guillaume Lample, Ludovic Denoyer, and Marc'Aurelio Ranzato. 2017.
\newblock Unsupervised machine translation using monolingual corpora only.

\bibitem[{Lample et~al.(2018)Lample, Ott, Conneau, Denoyer, and
  Ranzato}]{lample2018phrase}
Guillaume Lample, Myle Ott, Alexis Conneau, Ludovic Denoyer, and Marc'Aurelio
  Ranzato. 2018.
\newblock Phrase-based \& neural unsupervised machine translation.
\newblock \emph{arXiv preprint arXiv:1804.07755}.

\bibitem[{Lewis(2010)}]{lewis10haitian}
William~D. Lewis. 2010.
\newblock {Haitian Creole:} how to build and ship an {MT} engine from scratch
  in 4 days, 17 hours, \& 30 minutes.
\newblock In \emph{Proc. EAMT}.

\bibitem[{Malaviya et~al.(2017)Malaviya, Neubig, and Littell}]{malaviya17emnlp}
Chaitanya Malaviya, Graham Neubig, and Patrick Littell. 2017.
\newblock Learning language representations for typology prediction.
\newblock In \emph{Proc. EMNLP}.

\bibitem[{Munro(2010)}]{munro10crowdsourced}
Robert Munro. 2010.
\newblock Crowdsourced translation for emergency response in {Haiti}: the
  global collaboration of local knowledge.
\newblock In \emph{Proc. AMTA Workshop on Collaborative Crowdsourcing for
  Translation}.

\bibitem[{Neubig et~al.(2011)Neubig, Matsubayashi, Hagiwara, and
  Murakami}]{anpinlp11ijcnlp}
Graham Neubig, Yuichiroh Matsubayashi, Masato Hagiwara, and Koji Murakami.
  2011.
\newblock Safety information mining - what can {NLP} do in a disaster -.
\newblock In \emph{Proc. IJCNLP}, pages 965--973.

\bibitem[{Neubig et~al.(2018)Neubig, Sperber, Wang, Felix, Matthews,
  Padmanabhan, Qi, Sachan, Arthur, Godard, Hewitt, Riad, and
  Wang}]{neubig18xnmt}
Graham Neubig, Matthias Sperber, Xinyi Wang, Matthieu Felix, Austin Matthews,
  Sarguna Padmanabhan, Ye~Qi, Devendra~Singh Sachan, Philip Arthur, Pierre
  Godard, John Hewitt, Rachid Riad, and Liming Wang. 2018.
\newblock {XNMT}: The extensible neural machine translation toolkit.
\newblock In \emph{Proc. AMTA}, Boston.

\bibitem[{Nguyen and Chiang(2017)}]{nguyen-chiang:2017:I17-2}
Toan~Q. Nguyen and David Chiang. 2017.
\newblock Transfer learning across low-resource, related languages for neural
  machine translation.
\newblock pages 296--301.

\bibitem[{Qi et~al.(2018)Qi, Sachan, Felix, Padmanabhan, and
  Neubig}]{qi18naacl}
Ye~Qi, Devendra Sachan, Matthieu Felix, Sarguna Padmanabhan, and Graham Neubig.
  2018.
\newblock When and why are pre-trained word embeddings useful for neural
  machine translation?
\newblock In \emph{Proc. NAACL}, New Orleans, USA.

\bibitem[{Schultz and Black(2006)}]{schultz2006challenges}
Tanja Schultz and Alan~W Black. 2006.
\newblock Challenges with rapid adaptation of speech translation systems to new
  language pairs.
\newblock In \emph{Proc. ICASSP}, volume~5. IEEE.

\bibitem[{Sennrich et~al.(2016)Sennrich, Haddow, and Birch}]{sennrich16bpe}
Rico Sennrich, Barry Haddow, and Alexandra Birch. 2016.
\newblock Neural machine translation of rare words with subword units.
\newblock In \emph{Proc. ACL}, pages 1715--1725.

\bibitem[{Starbird et~al.(2012)Starbird, Muzny, and
  Palen}]{starbird12ontheground}
Kate Starbird, Grace Muzny, and Leysia Palen. 2012.
\newblock Learning from the crowd: Collaborative filtering techniques for
  identifying on-the-ground {Twitterers} during mass disruptions.
\newblock In \emph{Proc. ISCRAM}.

\bibitem[{Tiedemann(2018)}]{tiedemann2018emerging}
J{\"o}rg Tiedemann. 2018.
\newblock Emerging language spaces learned from massively multilingual corpora.
\newblock \emph{arXiv preprint arXiv:1802.00273}.

\bibitem[{Vieweg et~al.(2010)Vieweg, Hughes, Starbird, and
  Palen}]{vieweg10situational}
Sarah Vieweg, Amanda~L Hughes, Kate Starbird, and Leysia Palen. 2010.
\newblock Microblogging during two natural hazards events: what {Twitter} may
  contribute to situational awareness.
\newblock In \emph{Proc. CHI}, pages 1079--1088.

\bibitem[{Zoph et~al.(2016)Zoph, Yuret, May, and Knight}]{zoph2016transfer}
Barret Zoph, Deniz Yuret, Jonathan May, and Kevin Knight. 2016.
\newblock Transfer learning for low-resource neural machine translation.
\newblock In \emph{Proc. EMNLP}, pages 1568--1575.

\end{thebibliography}
\bibliographystyle{acl_natbib_nourl}

\end{document}